\documentclass{article}

\usepackage{arxiv}

\usepackage[utf8]{inputenc} 
\usepackage[T1]{fontenc}    
\usepackage{hyperref}       
\usepackage{url}            
\usepackage{booktabs}       
\usepackage{amsfonts}       
\usepackage{nicefrac}       
\usepackage{microtype}      
\usepackage{lipsum}
\usepackage{graphicx}
\usepackage{amsmath}
\graphicspath{ {./images/} }

\title{Coastline extraction from ALOS-2 satellite SAR images}

\author{
 Petr Hurtik \\
  IRAFM\\
  University of Ostrava\\ 
  30. dubna 22, Ostrava\\
  Czech Republic\\
  \texttt{petr.hurtik@osu.cz} \\
   \And
 Marek Vajgl \\
  IRAFM\\
  University of Ostrava\\ 
  30. dubna 22, Ostrava\\
  Czech Republic\\
  \texttt{marek.vajgl@osu.cz} \\
}

\begin{document}
\maketitle
\begin{abstract}
The continuous monitoring of a shore plays an essential role in designing strategies for shore protection against erosion. To avoid the effect of clouds and sunlight, satellite-based imagery with synthetic aperture radar is used to provide the required data. We show how such data can be processed using state-of-the-art methods, namely, by a deep-learning-based approach, to detect the coastline location. We split the process into data reading, data preprocessing, model training, inference, ensembling, and postprocessing, and describe the best techniques for each of the parts. Finally, we present our own solution that is able to precisely extract the coastline from an image even if it is not recognizable by a human. Our solution has been validated against the real GPS location of the coastline during Signate's competition, where it was runner-up among 109 teams across the whole world.
\end{abstract}


\section{Introduction and problem formulation}
As a country surrounded by oceans and a large length of the seaside, Japan is very sensitive to its changes, as every degradation is irreversible and affects the future. As the coast is under the continuous influence of water, it is affected by erosion. As mentioned in \cite{udo2017projections} and \cite{mori2018projection}, the erosion is accelerated by climate change, but also due to rapid national development. Such changes significantly affect the nearby population and property. To prevent the coast erosion, it is necessary to know where the store starts degrading. To monitor the coast, a standard imaginary does not propose sufficient performance because, regardless of its high resolution, it is very predisposed to artifacts caused by clouds and sunlight. The alternative source is satellite-based synthetic aperture radar -- SAR, which provides image data unaffected by clouds or sun. However, the quality of the provided SAR data is affected by selecting proper polarization, band, incident angle, and their processing demands using more advanced methods.

\begin{figure}[!h]
\centering
\includegraphics[width=45mm]{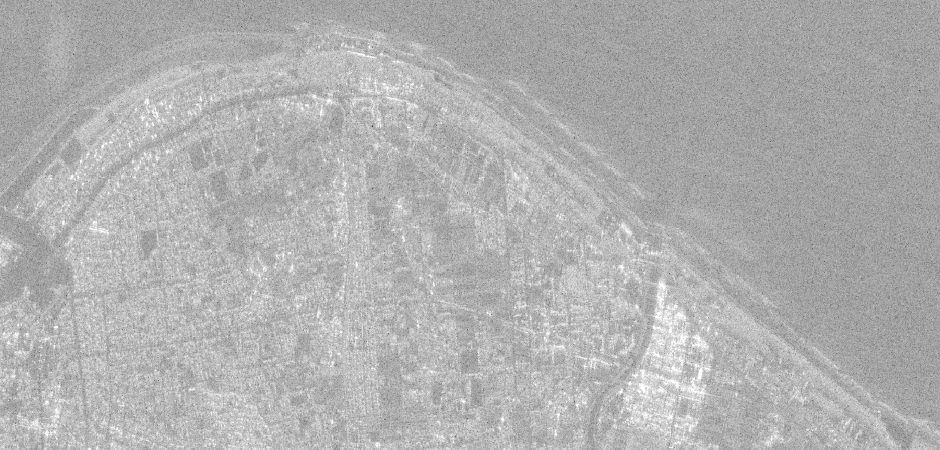}
\includegraphics[width=45mm]{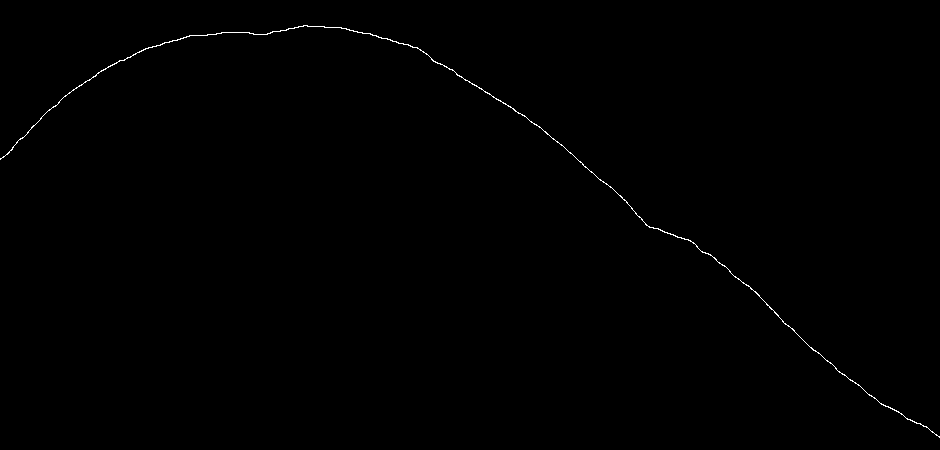}
\includegraphics[width=45mm]{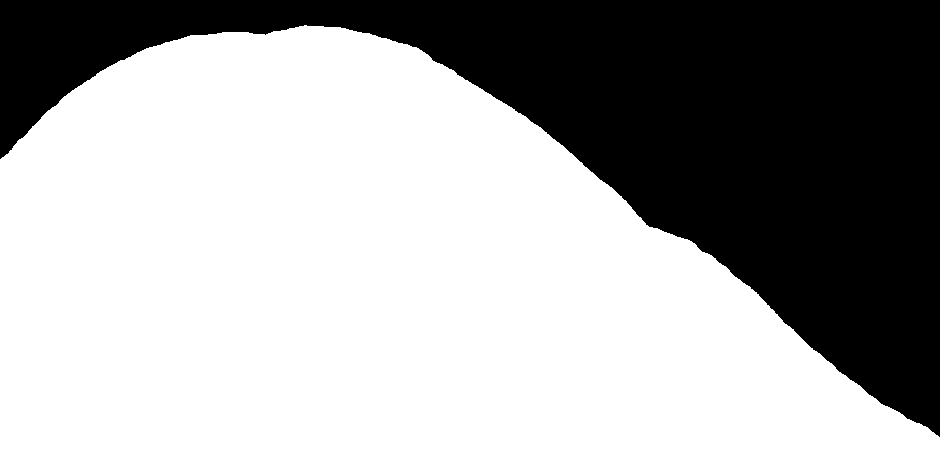}
\caption{The figure illustrates an image obtained from Tellus ALOS-2 (left, enhanced by log transform), corresponding ground-truth coastline (middle), and separation to sea and land area (right). Image credit: Tellus.} 
\label{fig-data}
\end{figure}

To tackle the problem and to exhaustively benchmark the real SOTA approaches in the area, the detection of coastline was the subject of an open, worldwide competition organized by Signate web site, where the attraction was supported by prize money. In this paper, we describe our pipeline and ideas used in the coastline detection competition\footnote{https://signate.jp/competitions/284}, where it achieved $2^{nd}$ place. The competition was sponsored by the Japanese Ministry of Economy, Trade, and Industry (METI), in cooperation with Japan Aerospace eXploration Agency. In total, 803 people registered for the competition, 147 of them (some of them in teams) submitted at least one solution. Every such submission was immediately evaluated on a partial test dataset to provide a preliminary ranking during the competition. From those submissions, every person/team selected one final submission, which (after the competition deadline) provides the final ranking on the full test dataset. In total, 2783 solutions were submitted. The competition lasted for three months, ending in November 2020. The training set included 25 high-res images with a total 461Mpx, and the testing set included 30 images with 777Mpx in total. The usage of external data was forbidden. The images were captured by ALOS-2 (JAXA) satellite with HH-polarization and the resolution of $3\times3$ meters. Unlike the 'standard' coastline extraction, the ground truth was prepared using GPS location instead of the visual observation using the captured images (for illustration, see Fig~\ref{fig-data}). That means the coastline can visually lie off the border observable from the images because such a border depends on the tide/ebb. It made the problem much more difficult. The evaluation computed the average Euclidean distance between the ground truth and the predicted coastline at predefined (unknown for the test set) evaluation points. If the algorithm missed the prediction, it was penalized. Our solution achieved the final standing score of 11.23; the first place achieved 11.18, and the third place reached a score of 11.76. For details how the score was computed, see signate.jp/competitions/284\#evaluation.

\begin{figure}[!h]
\centering
\includegraphics[width=140mm]{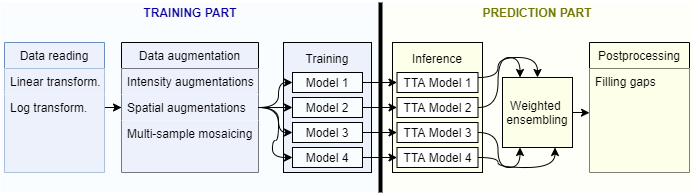}
\caption{The image shows the pipeline for coastline extraction, split into the training and prediction part. A detailed description is the subject of this study.} 
\label{fig-pipeline}
\end{figure}

Our pipeline include existing augmentations as well as one novel, powerful architecture with modern SOTA backbones and powerful postprocessing. As our solution is based on a data-driven approach, it can be applied to other remote sensing problems where image segmentation is needed. All source codes are available online\footnote{https://gitlab.com/irafm-ai/signate\_4th\_tellus\_competition} with a copyleft license. The full pipeline is shown in Figure~\ref{fig-pipeline}.

The contribution and the paper's goal is to create a 'cookbook' of good practices and solutions for difficulties in remote sensing image segmentation. The novelty is in the multi-sample mosaicing augmentation, which increases the model's generalization ability and leads to higher accuracy.

\section{Current state}

\textbf{The model-driven approaches}, say the classical ones, usually involve methods known from standard image processing such as kernel-based filtering, morphology, or gradient operators. Such a typical example is Acar's et al. approach~\cite{acar2012algorithm} that filters out the image by histogram equalization and mathematical morphology until Sobel's operator can extract coastline. Wiele et al.~\cite{wiehle2015waterline,wiehle2015automated} improved the process of iterative gap closing and flood filling. Mara et al. ~\cite{maracs2016automatic} use median filtering to homogenize an image, which is then input for segmentation and morphological edge extractor. Mirsane et al. ~\cite{mirsane2018automatic} use a wavelet algorithm followed by iso-data extraction, filled by a watershed algorithm and take the coastline as a border from the largest segment in the image. The obtained shape is smoothened by mathematical morphology. Demir et al.~\cite{demir2016extraction} realize fuzzy-based classification of pixels to land/sea classes and extract the coastline from it.

The disadvantages of model-driven approaches are possible large computation time~\cite{maracs2016automatic} and mainly low generalization -- they fail when the input's parameters are changed. That is caused because the model-driven approaches usually have fixed spatial and intensity dependencies by a bunch of hand-tuned parameters~\cite{demir2016extraction}.

\textbf{In the case of data-driven approaches}, we can distinguish between those based on Deep Neural Networks (DNN) and the others. The others involve optimization algorithms, for example, Particle swarm optimization~\cite{reis2018extended}. These optimization methods did not prove their usefulness and, therefore, are used only rarely.

Shamsolmoali et al. \cite{shamsolmoali2019novel} propose to use a residual U-Net~\cite{zhang2018road} architecture to segment an input image into sea/land regions. From that, the coastline can be extracted by the way described in model-driven approaches.  Kesikoglu et al. \cite{kesikoglu2017determination} works with an obsolete fully-connected architecture, but proposes a pipeline that can directly produce changes in the coastline for images with various time stamps. Dang et al. \cite{dang2020convolutional} use a light architecture for coastline classification consisting of five layers only, where the most trainable parameters are in the two last fully connected layers. The lightweight architecture allows the usage of relatively high input resolution.

The approaches based on DNNs suffered mainly for big training datasets' requirements. This paper shows that with a proper pipeline where pre/postprocessing is carefully designed, it is possible to use even small training datasets, namely, 25 images in our case.

\section{Preparing data}
For training deep learning models, it is necessary to prepare the data in a suitable format. Here, we firstly describe how the data are loaded and transformed, and secondly, how the data are augmented, i.e., distorted to increase the model's generalization ability.

\subsection{Handling the images}
The provided images are in tiff format with a pixel value range in [0,65535]. There are two ways to preprocess such images into the $[0,1]$ range, commonly used in deep learning. The first one is based on a linear mapping, i.e., $f'=f/65535$, where $f$ is the original image and $f'$ is the output one. The second one reflects the physical properties of satellites (noise ratio) and it is given a non-linear transformation $f' = 10\log(f^2+c)$, where $c$ is a noise reduction coefficient, namely $c=-83$ in our case \cite{wu2019study}. There is a general knowledge that the input of a neural network should be the rawest, and the network will select such data/features itself during the training process. Therefore, the linear mapping should be selected, but, in our experiments, the non-linear-input-based models yield slightly better accuracy. Therefore, we used an ensemble of models where we applied both models based on linear and non-linear inputs.

The additional manipulation can involve image rescale/crop according to the model resolution. There are two main reasons to use cropping and avoid rescaling: firstly, the training dataset is small; it has 25 images. Thus, there is a need to use as much information as possible. If we rescale the images to the resolution of the model (defined in Section~\ref{model}), for some cases, we use less than one percent of the original data. Secondly, all images have the same physical resolution, i.e., one pixel in the image always represents the same size in meters, but they have different resolutions, i.e., they capture a different portion of the landscape. Their rescale to the same resolution invokes additional distortion. Based on the two reasons, the rescaling yields poor performance compared to the original data usage.
Regarding the cropping, if we have an image with the resolution of $w\times h$ and crop resolution of $a\times b$, we can create $(w-a)(h-b)$ unique overlapping crops. Considering a model resolution of $512\times 512$px and the resolution of the particular images, we can obtain more than 365M unique crops. Because it is impossible to save such an amount on a hard drive, we did not realize the cropping in advance and postponed it to augmentations performed during the training.

The labels are as follows. They include values of $\{0,1\}$: the pixel has value 1 if it is a coastline, 0 otherwise. The classes are highly imbalanced, i.e., the total area of the coastline is tiny compared to the rest. To postpone the problem of such imbalance, we applied a form of label smoothing technique~\cite{muller2019does} and obtained labels in the interval [0,1]. We also created a second form of labels. We manually notated them as sea / land / no-data. The no-data class is required because parts of some images have missing information. Such labels are more balanced and include more information. In our pipeline, some models were trained on the original labels (coastline), and some were trained on the modified version to secure the diversity of the models in the ensemble. For details, see Table~\ref{tab-configuration}.

\subsection{Data augmentation}
\label{subsec_data_augmentation}
\begin{figure}[!h]
\centering
\includegraphics[height=19.5mm]{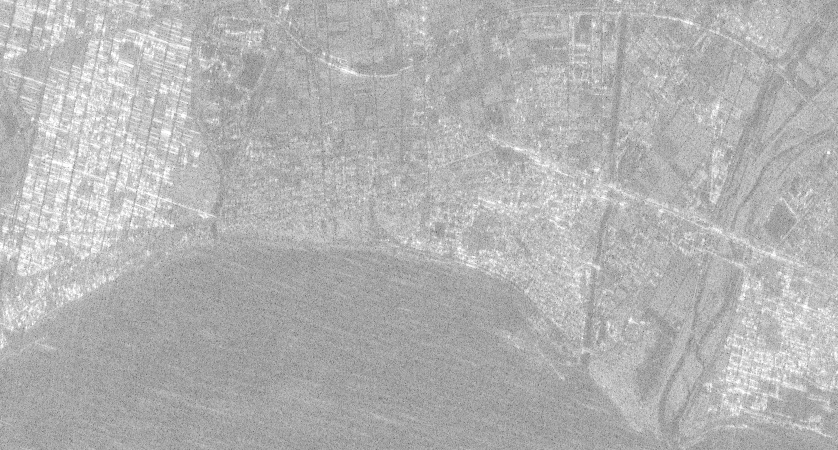}
\includegraphics[height=19.5mm]{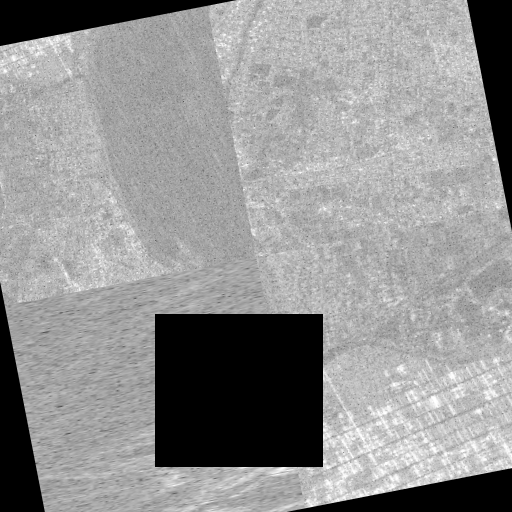}
\includegraphics[height=19.5mm]{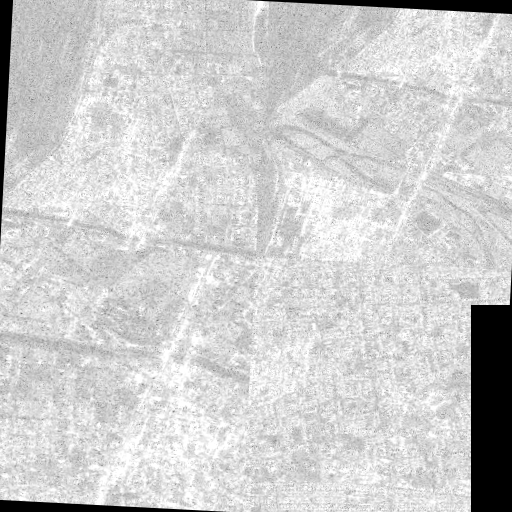}
\includegraphics[height=19.5mm]{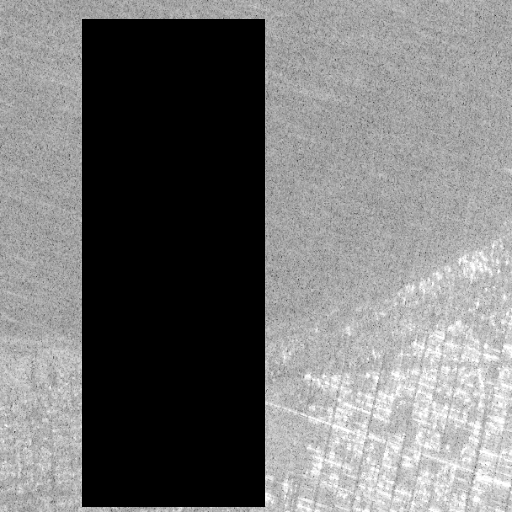}
\includegraphics[height=19.5mm]{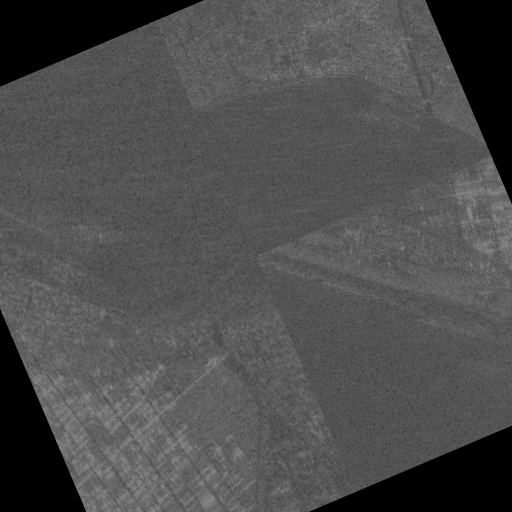}
\includegraphics[height=19.5mm]{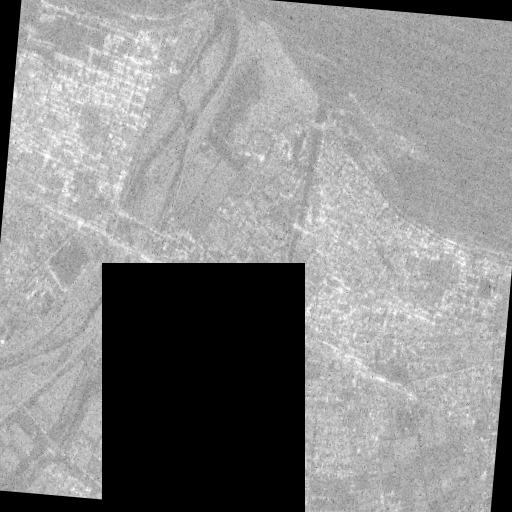}
\\[1mm]
\includegraphics[height=19.5mm]{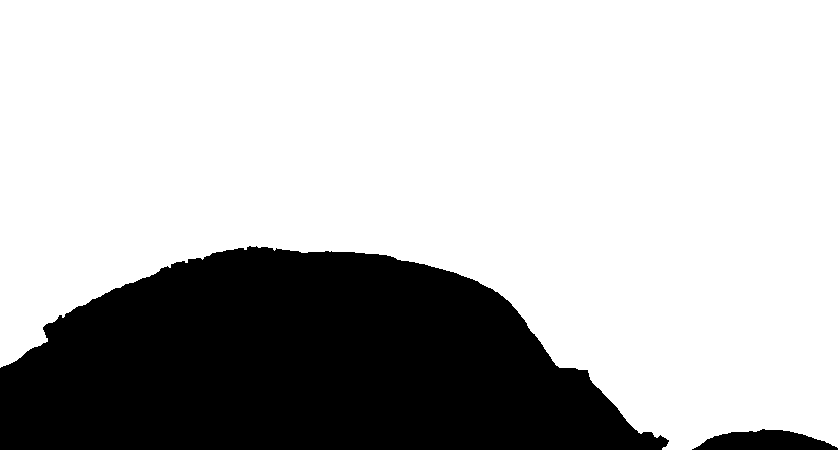}
\includegraphics[height=19.5mm]{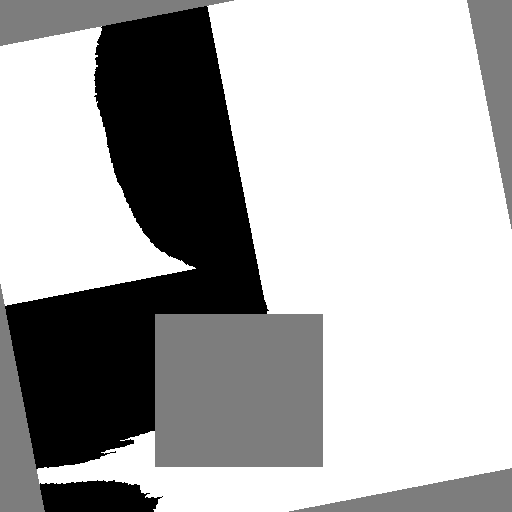}
\includegraphics[height=19.5mm]{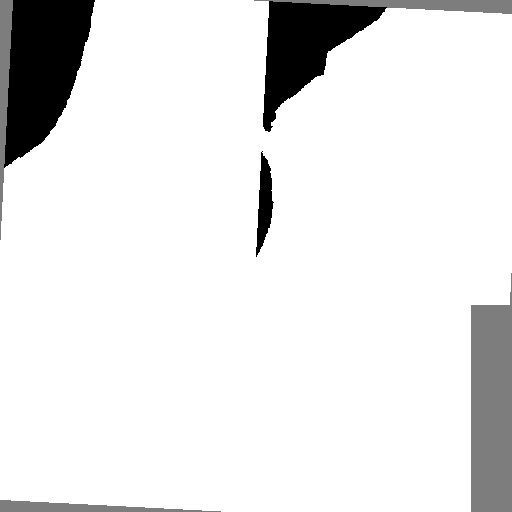}
\includegraphics[height=19.5mm]{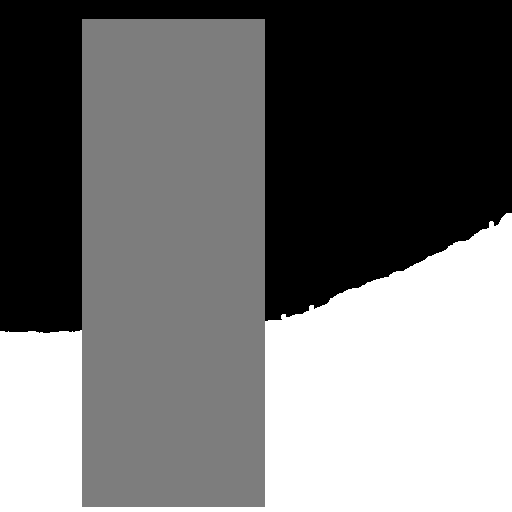}
\includegraphics[height=19.5mm]{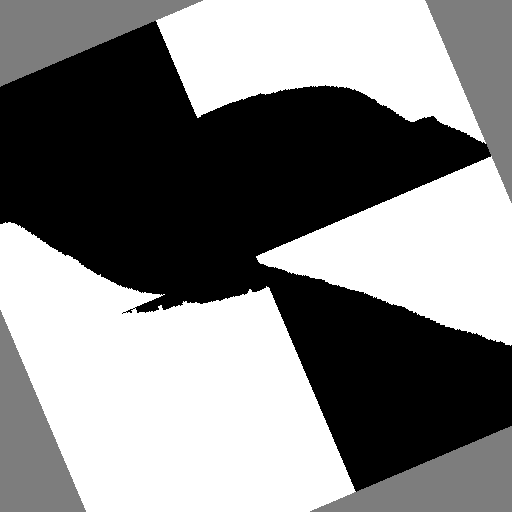}
\includegraphics[height=19.5mm]{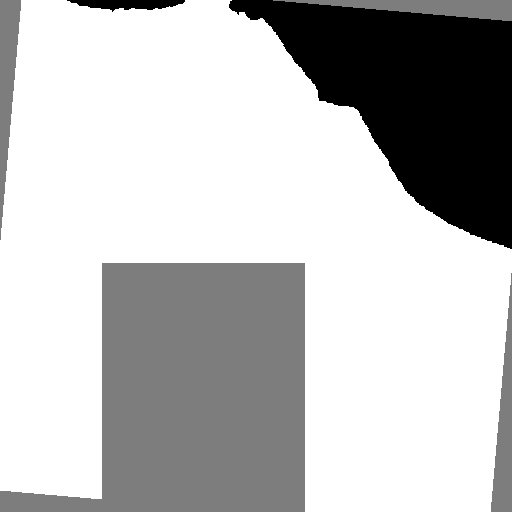}
\caption{The left images show an image from the dataset (top) with its label (bottom) in full resolution. The other images are crops created from the original involving the augmentations described in Section~\ref{subsec_data_augmentation}. With the augmentations, we can create uncountable crops. The original image credit: Tellus.} 
\label{fig-augment}
\end{figure}

The role of data augmentation is to create new image samples, which help to reduce the model's overfitting. It is realized during the training and can be generally divided into intensity and spatial augmentation. Intensity augmentation involves additive, multiplicative, or gamma intensity shifts, additive or multiplicative noise, blurring, or Cropout~\cite{hou2019cropout}. The parameters of augmentation should always reflect the manner of the data; it is useless to create entirely artificial samples that cannot be met during the inference.
The spatial augmentation includes flips, rotation, rescale, creating random crops, elastic transformations, or mosaicing. In our pipeline, we took a crop at a random position in the image with a side-size in the interval [1024,1536] and downscaled it to our model's side-size, 512. Then we applied other spatial augmentations. Finally, we realized our own augmentation, \textbf{multi-sample mosaicing}. It means that we split the data sample into $n$ rectangular parts and replaced some of them with a same-sized data area from a different image from the training set. The main advantage is that such a composed image includes multiple characteristics, simulating a bigger batch size and, therefore, postpones overfitting. An example of the input image and its augmented versions are shown in Figure~\ref{fig-augment}.

Note, there is a well-established library called Albumentations~\cite{buslaev2020albumentations} realizing the augmentations. Because we involved our own augmentation, multi-sample mosaicing, we created custom functionality and did not use Albumentations.

\section{Architecture and training of the derived model}
\label{model}
In this section, we discuss suitable neural network architectures for image segmentation as well as the loss functions and optimizer. The selected combination(s) is then the subject of training, i.e., searching for the optimal setting of the model's weights to minimize the loss function value. During the training, the model takes the augmented data iteratively. The output of the training process is a model that is used for prediction.

The segmentation architectures are based on an encoder-decoder scheme, i.e., it decodes the input representation into a latent one, which is then projected back (decoded) to the original resolution. The spatial dimension is reduced during the encoding while the feature dimension is increased, and vice-versa for the decoding phase. The typical representative is U-Net~\cite{ronneberger2015u} which has been evolved into a residual form or a nested version~\cite{zhou2018unet}.  The advantage of U-Net is the simplicity of coding and low memory requirements compared with the other architectures. The others are, e.g., LinkNet~\cite{chaurasia2017linknet} or Feature Pyramid Network~\cite{lin2017feature}, which is then also used in DeepLabv3~\cite{chen2018encoder} that utilizes atrous convolution. We have selected U-Net in the competition because it allows us to use a bigger batch size than other architectures with the same setting and generally holds that a bigger batch size means lower overfitting. Compared to FPN, it also yields a lower error, namely, 16.2 vs. 17.2. Note, the absolute winner of the Signate's competition reported that he used U-Net as well.

Because the encoder's (in U-Net) ability to extract features is limited, it is beneficial to replace the default encoder with some of the SOTA networks known, e.g., from the classification problem. These networks are called backbones and can be pre-trained on ImageNet~\cite{russakovsky2015imagenet} to converge faster. The most powerful (see the comparison in~\cite{bianco2018benchmark}) are ResNet, SENet, ResNeXt, Inception-ResNet, RegNet, or Xception, to name a few. In our pipeline, we have selected EfficientNet~\cite{tan2019efficientnet}, or EffNet for short. It is a fully convolutional architecture based on compound scaling that allows easily controlling the trade-off between the network capacity and memory requirements. The detailed configuration is given in Table~\ref{tab-configuration}.


\begin{table}[!h]
\caption{Configuration of the models}
{\begin{tabular}{llllll}\toprule
 Backbone&Input&Classes& Loss & Last layer\\ 
 \midrule
 EffNetB4 & log   & sea/land/no-data/coast & Dice+Focal & Softmax\\
 EffNetB4 & log   & sea/land/no-data & Dice+Focal & Softmax\\
 EffNetB3 & log   & coast & BCE & Sigmoid\\
 EffNetB3 & linear& coast & BCE & Sigmoid\\
 \bottomrule
\end{tabular}}
\label{tab-configuration}
\end{table}

Because we planned to create an ensemble of different models, we trained several models based on U-Net with EffNet backbone. Regarding the loss function, the commons are based on group characteristics such as IOU, Tversky loss, or Sorensen-Dice loss~\cite{wang2020improved}, or pixel characteristics, like Binary cross-entropy (BCE) or Focal loss~\cite{lin2017focal}. Each of them has pros and cons. Sorensen-Dice considers the spatial dimension but generally does not lead to the best performance; Focal loss can partially solve the class imbalance problem but may overfit; BCE can be marked as a good and universal baseline. In our pipeline, we combine Dice with Focal for the two models and use BCE for the other two models, see Table~\ref{tab-configuration}. 

Regarding the optimizer, the first choice is generally Adam \cite{kingma2014adam}, a combination of AdaGrad and RMSProp, which has been several times marked as one of the best optimizers. On the other hand, there are known datasets (such as CIFAR-10 classification) where it yields sub-optimal performance. In our experiment~\cite{vajgl2020pipeline}, we have confirmed the behavior. Therefore, we used Adam optimizer for the two models and AdaDelta~\cite{zeiler2012adadelta} for the next two.

The rest of the training setting is as follows. We use the models' resolution equal to $512\times512$px and as big batch size as possible, varying from 3 to 12 according to the used GPU. The models were trained for 100 epochs with reducing the learning rate on a plateau and with saving the best model according to the validation dataset. For more details, see our repository. The models with sea/land/no-data labels have in the last layer softmax (a smooth approximation of one-hot argmax); the models with the coastline class only have in the last layer sigmoid (a smooth logistic function). It means the former creates a decision between the classes, and the latter produces the probability of being a coastline. The training of the four models takes approx three days on an RTX2080Ti graphics card.

\section{Inference}
In the inference part, we take the trained models, read the testing images, and produce the final predicted coastline.  The pipeline is as follows: each model firstly realizes prediction on the basis of the test time augmentation principle (TTA). From the predictions, we extract the coastlines. Then, the four coastlines (produced by four models) are aggregated into one. Finally, we apply postprocessing for filling gaps in the coastline.

\textbf{To produce predictions}, each model creates its own ensemble. We use the technique of floating window, where we create overlapping crops in multi-scale resolutions equal to the conditions we had during the training phase. Because the inference is significantly less demanding for memory than training, we are able to process hundreds of crops at once, so the calculation is fast. When the predictions are projected back into the original image, the overlapping parts can be aggregated by summation because the process of extracting the coastline described above does not depend on the absolute values. A Gaussian filter smoothes these produced predictions to decrease the impact of noisy outliers.

\textbf{The process of extracting the coastline} differs for models with softmax in the last layer and for models with sigmoid in the last layer. 

\emph{Softmax models:} These models use the following label-into-class encoding: 0=sea, 1=no-data, 2=land. The models produce a structure $f$ where each pixel $(x,y) \in f$ is a vector of three values with a probability of beeing a certain class. Firstly, we create $f'(x, y) = \text{argmax}(f(x,y))$, so it holds that $f'(x,y) \in \{0,1,2\}$. From it, we create the final 'coastline' image $f''$ as 
$
f''(x,y) =
\begin{cases}
    1, & \Delta_{max}(x,y) - \Delta_{min}(x,y)  = 2\\
    0,              & \text{otherwise}
\end{cases},
$
where $\Delta_{max}$ and $\Delta_{min}$ extract the maximum and minimum values of $f'$ in a $3\times 3$ neighborhood of $(x,y)$. For $f''$ holds that $f''(x,y)=1$ marks the presence of coastline and $f''(x,y)=0$ no coastline. In other words, we say that there is a coastline if some area contains both 'sea' and 'land' classes without taking into account the 'no-data' class.

\emph{Sigmoid models:} Firstly, we initialize $f''(x, y) = 0$ for all $(x,y)$, and then check if the image has a landscape or portrait orientation. For the landscape, we browse all $x$ coordinates and for each of them we set $f''(x, \text{argmax}(f(x,\cdot)) = 1$ whereby $\cdot$ means all $y$ coordinates. In other words, we are searching for the maximum probability of being a coastline in each column for all rows. The process is the same for the landscape, but we search for each column's maximum in a row. The advantage of the process is the absence of a threshold, so we are able to extract even the most uncertain coastlines. The disadvantage is that we can miss coastline points if the coastline slope is stronger than the diagonal, or miss a coastline if there are two coastlines in a row/column. We suppress the disadvantages in the postprocessing later.

\begin{figure}[!h]
\centering
\includegraphics[width=100mm]{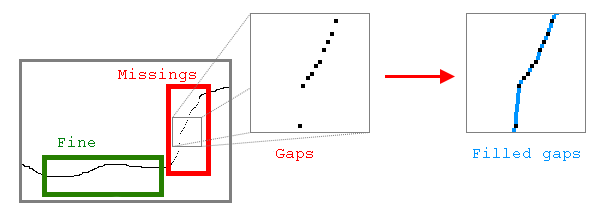}
\caption{Filling the gaps in the coastline detection. Black line represents the coastline. For horizontal coastline, vertical part (red rectangle) contains gaps, which are detected and filled (blue color).}
\label{fig-fill-gaps}
\end{figure}

\textbf{Ensemble of coastlines and postprocessing:} The output image functions $f''$ of the four models have been taken and processed in the following way. We browsed the images column/row-wise, as same as when we made predictions for models with sigmoid. If we browse the rows, then we find the coordinates of the coastline in the columns and create the final prediction as the weighted average of the four predictions. The models' weights have been set according to a particular model evaluation in the public competition's leaderboard. Because the realized way of extracting coastlines can create holes when the coastline's slope is big, we apply postprocessing, which searches for such holes and fills them. The illustration of the process is shown in Figure~\ref{fig-fill-gaps}.

\section{Overall results and sumarization}
\begin{figure}[!h]
\centering
\includegraphics[height=36mm]{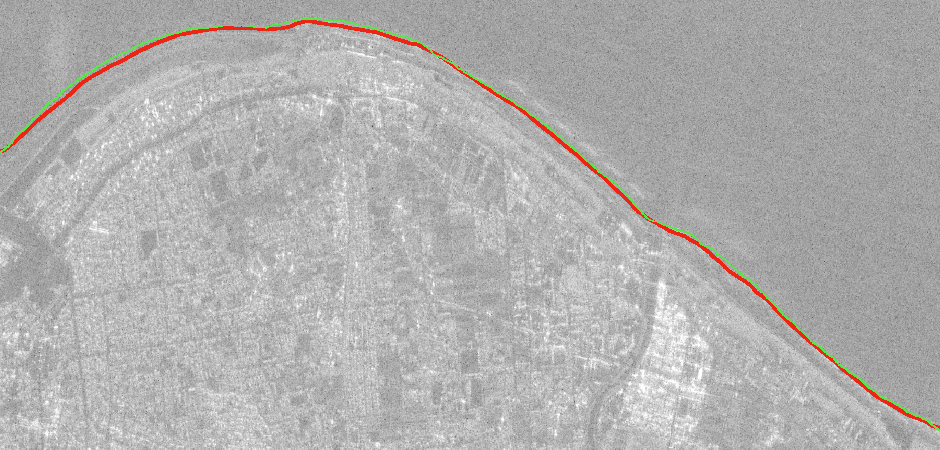}
\includegraphics[height=36mm]{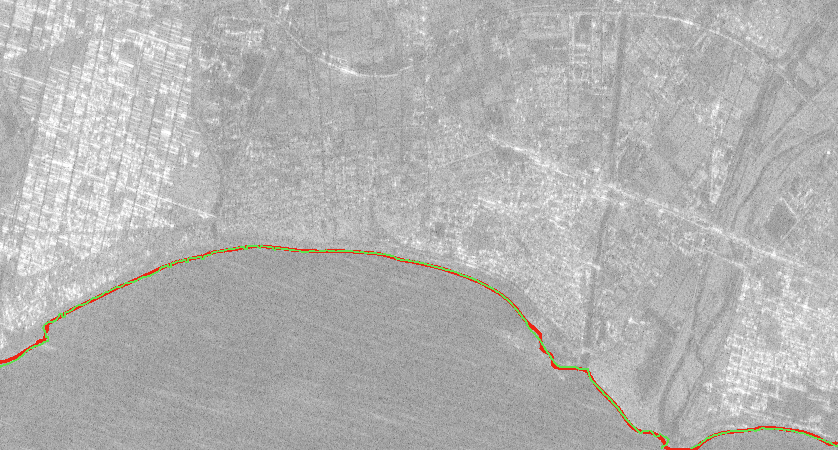}
\caption{Here we show two images (with log transform), where our model realized inference and inpainted the detected coastline with red color. The green color denotes the real coastline measured using GPS. It is obvious that standard algorithms based on edges or contours will fail because the real coastline lies in shape without a significantly visible gradient. The original image credit: Tellus.} 
\label{fig-data-vysledek}
\end{figure}

In this paper, we deal with the problem of coastline extraction from SAR images. We have recalled the current state and justified the importance of the deep-learning-based approach. We have presented a pipeline split into data preprocessing and augmentation, architecture selection, training of a model, and inference with ensembling and postprocessing. All presented pieces of the mosaic are carefully selected and discussed in the state-of-the-art context. The measured impact of various aspects of hyperparameters regarding the competition's score is shown in Table~\ref{tab-difference}. The described solution forms a general pipeline that can be applied to many segmentation tasks and whose superiority has been confirmed by reaching the second place in world-wide Signate competition. For the visual performance, see Figure~\ref{fig-data-vysledek}.
The proposed pipeline is modular, which means it can be easily updated by more data, new augmentation types, or more powerful backbones.

\begin{table}[!h]
\caption{The table shows the ablation study where we measured the impact of various aspects on the official competition score.}
{\begin{tabular}{lr}\toprule 
 Description &Improvement\\ 
 \midrule
 Input resolution of the model 512x512px instead of 256x256px &  0.48\\
 Used architecture U-Net instead of FPN	&	1.04\\
 Model's backbone EfficientNetB3 instead of SE-ResNeXt50	&	1.19\\
 Models's Backbone EfficientNetB6 instead of EfficientNetB3  &	-1.99\\
Applied histogram adjustment as preprocessing	&	-3.12\\
Softmax activation in the last layer instead of Sigmoid in 'sea/land/no-data' model	&	1.08\\
Adding multi-sample mosaicing to data augmentation	&	0.52\\
Curriculum learning of the models	&	0.44\\
TTA realized by overlapping crops	&	0.62\\
TTA improved by multiscale crops &	0.04\\
Gaussian smoohing before coastline extraction is realized	&	0.14\\
Ensemble of four models	&	1.41\\
Postprocessing, filled gaps in coastline	&	0.42\\
\bottomrule
\end{tabular}}
\label{tab-difference}
\end{table}

\section*{Acknowledgment}
We want to thank Dr. Kunihiro Watanabe (National Institute for Land and Infrastructure Management, Japan) for his valuable pieces of advice and comments.

\bibliographystyle{unsrt}  
\bibliography{tellus}  

\end{document}